%% file: main.tex
\documentclass[10pt,twocolumn,letterpaper]{article}

\usepackage{iccv}
\usepackage{times}
\usepackage{epsfig}
\usepackage{graphicx}
\usepackage{amsmath}
\usepackage{amssymb}
\usepackage{booktabs}
\usepackage{bbm}
\usepackage{pifont}
\usepackage{enumitem}
\usepackage{balance}
\usepackage{colortbl}
\usepackage{multirow}
\usepackage[font=small,skip=4pt]{caption}
\usepackage[pagebackref=true,breaklinks=true,colorlinks,bookmarks=false]{hyperref}
\usepackage[capitalize]{cleveref}
\crefname{section}{Sec.}{Secs.}
\Crefname{section}{Section}{Sections}
\Crefname{table}{Table}{Tables}
\crefname{table}{Tab.}{Tabs.}
\input{math}
\newcommand{\tinyspace}{\vspace{0.3mm}}
\definecolor{Gray}{gray}{0.95}
\newcommand{\method}{ResQ~}
\newcommand{\dymethod}{Dynamic-ResQ~}
\newcommand{\basemethod}{Frame Quantization~}

\DeclareMathOperator*{\argmin}{arg\,min}
\newcommand{\amprec}[2]{#1$\rightarrow$#2}
\iccvfinalcopy

\ificcvfinal\fi
\begin{document}
\title{ResQ: Residual Quantization for Video Perception}
\author{
Davide Abati \qquad Haitam Ben Yahia \qquad Markus Nagel \qquad Amirhossein Habibian 
\vspace{0.5em}
\\
Qualcomm AI Research\thanks{Qualcomm AI Research is an initiative of Qualcomm Technologies, Inc.}\\
\small{\texttt{\{dabati,hyahia,markusn,ahabibia\}@qti.qualcomm.com}}
}

\maketitle
\ificcvfinal\thispagestyle{empty}\fi
\input{sections/0_abstract}
\input{sections/1_introduction}
\input{sections/2_related_work}
\input{sections/3_motivation}
\input{sections/4_method}
\input{sections/5_experiments}
\input{sections/6_conclusions}
\clearpage
\twocolumn[
\begin{center}
\Large
\textbf{
ResQ: Residual Quantization for Video Perception\\Supplementary Material
}
\end{center}
]
\input{sections/supp_text/additional_exps}
\input{sections/supp_text/latency}
\input{sections/supp_text/qualitative_results}
\clearpage
\balance
{\small
\bibliographystyle{ieee_fullname}
\bibliography{references}
}
\end{document}

%% file: math.tex
\def\dlt{\boldsymbol\delta}

\def\x{\mathbf{x}}
\def\xhat{\mathbf{\hat{x}}}
\def\w{\mathbf{w}}
\def\what{\mathbf{\hat{w}}}
\def\z{\mathbf{z}}
\def\zhat{\mathbf{\hat{z}}}

\def\X{\mathbf{X}}
\def\q{q}

\def\rmin{\text{r}_{min}}
\def\rmax{\text{r}_{max}}
\def\s{s}
\def\q{\mathbf{q}}

%% file: sections/0_abstract.tex
\begin{abstract}
This paper accelerates video perception, such as segmentation and human pose estimation, by levering cross-frame redundancies.
Unlike the existing approaches, which avoid redundant computations by warping the past features using optical-flow or by performing sparse convolutions on frame differences, we approach the problem from a different perspective: low-bit quantization.
We observe that residuals, as the difference in network activations between two neighboring frames, exhibit properties that make them highly quantizable.
Based on this observation, we propose a novel quantization scheme for video networks coined as \underline{Res}idual \underline{Q}uantization. ResQ extends the standard, frame-by-frame, quantization scheme by incorporating temporal dependencies that lead to better performance in terms of accuracy vs. bit-width.
Furthermore, we extend our model to dynamically adjust the bit-width proportionally to the amount of changes in the video.
We showcase the superiority of our model, against the standard quantization and existing efficient video perception models, using various architectures on semantic segmentation, video object segmentation and human pose estimation benchmarks.
\end{abstract}

%% file: sections/1_introduction.tex
\section{Introduction}
Despite the great progress made in optimizing the computational efficiency of deep neural networks, real-time video inference is still an open challenge in many cases, especially when deploying on resource-constrained devices~\cite{deltacnn,deltadist,zhu17dff,skipconv,eventnets,hu2020tdnet,accel}.
The most effective strategy to accelerate video inference is to exploit the temporal redundancy,~\ie by leveraging what has been processed already in the past.
Early works relied on feature warping and adaptation using optical flow~\cite{zhu17dff,accel} or self-attention~\cite{hu2020tdnet}.
More recently, some encouraging results have been obtained by decomposing video snippets into a keyframe followed by residuals, that can be more efficiently processed with distilled networks~\cite{deltadist} or sparse operators~\cite{skipconv,deltacnn,eventnets}.

For many applications, neural network quantization has emerged over the years as one of the key techniques for accelerating floating point models~\cite{lsq,nagel2019dfq,krish2018quant,adaround,bannerposttraining,jacob2018cvpr}, and for their deployment in integer precision.
However, when dealing with video inputs, quantized models still operate by processing every frame independently, neglecting the opportunity to exploit redundancy among frames.

\input{figures/cover}
In this paper, we observe that residuals, as the difference in network activations between adjacent frames, exhibit properties that make them highly quantizable.
Specifically, we illustrate that residuals typically exhibit a smaller variance \wrt the frame activations, which results in a reduction in quantization error as illustrated in~\cref{fig:cover}. 
Following this observation, we propose Residual Quantization, coined as \textit{ResQ}, a novel quantization scheme tailored for video perception.
ResQ employs two sets of quantizers: one at a higher precision to quantize keyframes, one at a lower precision to quantize the residuals for subsequent frames.
Both quantizers interact during the inference to combine the high-precision details from keyframes with complementary information from residuals.
Furthermore, motivated by the fact that the range of residuals depends on the scene, we extend \method to dynamically adjust the quantization bit-width on the fly.
More specifically, we propose a lightweight policy function that assigns a minimally acceptable bit-width when the residuals are small, such as in static scenes.

We extensively evaluate our proposals on three perception tasks, namely semantic segmentation, video object segmentation and human pose estimation.
We experiment with various architectures and quantization techniques, \ie post-training quantization and quantization-aware training.
\method and its dynamic counterpart consistently outperform the standard quantization schemes, and perform favorably compared to state-of-the-art in efficient video processing.

We summarize our contributions as follows:
\emph{i)}
We formally and empirically show the benefits of using frame residuals in reducing the quantization error (\cref{sec:motivation});
\emph{ii)}
We propose \method, a novel quantization scheme for video networks leveraging the residual quantization (\cref{sec:delta_quantization});
\emph{iii)}
We extend our model to dynamically adjust the quantization level based on the residual content (\cref{sec:dynamic});
\emph{iv)}
We validate our proposals on three tasks, where our models achieve an optimal trade-off of accuracy vs. efficiency.

%% file: figures/cover.tex
\bgroup
\setlength{\tabcolsep}{1.5pt}
\renewcommand{\arraystretch}{0.5}
\begin{figure}[t]
\centering
\resizebox{0.95\columnwidth}{!}{
\begin{tabular}{cc}
\includegraphics[width=0.5\columnwidth]{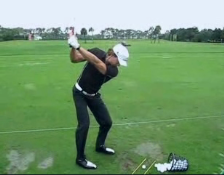}&
\includegraphics[width=0.5\columnwidth]{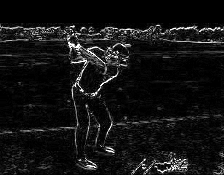}\\
\includegraphics[width=0.5\columnwidth]{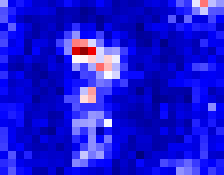}&
\includegraphics[width=0.5\columnwidth]{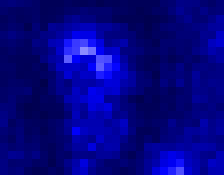}\\
Frame quantization. err. & Residual quantization. err.
\end{tabular}}
\caption{\textbf{Visualization of the quantization error} (bottom) when using frame representations (left) and residual representations (right). The residual quantization. error is typically lower (blue) than the one on individual frames (red).}
\label{fig:cover}
\end{figure}
\egroup

%% file: sections/2_related_work.tex
\section{Related Work}
\label{sec:related}
\paragraph{Neural network quantization.}
Progress in building efficient neural networks followed from multiple directions,~\eg by pruning parameters~\cite{denil2013PredictingPI,comp2,li2017PruningFF}, decomposing weights into low-rank tensors~\cite{comp2,zhang2016acceleratingvd}, searching for optimal architectures~\cite{fasterseg,donna} and distilling knowledge into light models~\cite{hinton,fitnets}
Besides these strategies, low-precision network quantization~\cite{Gupta2015,jacob2018cvpr,krish2018quant,nagel2019dfq,whitepaper} has proven to be extremely effective, mostly due to the robustness of deep models to noise in representations and to the wide support in the acceleration of fixed-point inference.
In Post-Training Quantization (PTQ), a trained neural network is quantized without fine-tuning its parameters.
This can be achieved by optimizing the quantization ranges~\cite{krish2018quant, bannerposttraining}, correcting biases in quantization error~\cite{nagel2019dfq, biaswithbias}, performing layer-wise rescaling~\cite{nagel2019dfq, samesame}, or using sophisticated rounding operations~\cite{adaround, brecq}.
Whenever labeled data is available it is possible to perform Quantization-Aware Training (QAT), to fine-tune the model while simulating quantized operations.
In this respect, the gradient of rounding operations can be approximated by the straight-through estimator (STE)~\cite{straightthrough, Gupta2015, jacob2018cvpr} or its variations~\cite{DSQ, EWGS, PBGS}.
For lower bit-widths, it is beneficial to learn the quantization ranges through gradient decent~\cite{lsq, pact2018, tqt}, possibly jointly with bin regularization~\cite{binreg} or 
oscillation damping~\cite{nagel2022oscillations}.
Unlike all these efforts, we leverage temporal redundancies, a novel and complementary aspect to improve the quantization of models when being applied on a sequence of frames.
\paragraph{Efficient video perception.}
Most relevant to our effort are the works that leverage temporal redundancies to accelerate video processing. 
The seminal work Deep feature flow (DFF)~\cite{zhu17dff} and its extensions~\cite{accel,flow_guided,li2018low,liu2018mobile} avoid expensive feature computation by warping features from key-frames using optical-flow.
Despite its effectiveness for expensive backbones, feature warping has become less appealing recently, as the cost of optical-flow estimation might outweigh feature computation.
In TDNet~\cite{hu2020tdnet} a wide backbone is divided into several sub-models, to whom frames are cyclically assigned;
current and past representations are then merged via self-attention.
Differently, our method saves computation by reducing quantization precision rather than network width, and does not require any additional temporal modeling or other modifications to the original architecture.
Recent works focus on representing videos as residuals w.r.t. keyframes~\cite{skipconv, deltadist, deltacnn, eventnets}.
Skip-convolutions~\cite{skipconv} showcase how dense features can be approximated by performing few localized operations over residuals, leading to impressive theoretical speed-ups.
However, their dependency on sparse convolutions makes them hard to deploy in practice.
Despite more recent efforts~\cite{deltacnn, eventnets, deltadist}, the development of practical solutions to process residuals is still an open challenge.
\\\\
Despite the potential impact, no prior work explored how to leverage correlation of video frames to improve neural network quantization.
A notable exception is
VideoIQ~\cite{sun2021dynamic}, dedicated for action recognition. 
Specifically, it dynamically quantizes frames at high or low precision based on their importance to classify a video: higher bit-widths for \textit{foreground} frames and lower bit-widths for \textit{background} frames.
As we tackle per-frame prediction tasks,~\eg semantic segmentation and pose estimation, \method aims at a precise representation on \textit{all frames}.
This aim is achieved by leveraging frame residuals, which is missing in VideoIQ.
Moreover, VideoIQ relies on a complex recurrent policy model, trained jointly with the action classifier using discrete optimization and knowledge distillation.
By relying only on the residual representation, \method is simply applicable to any off-the-shelf model with minimal modifications.

%% file: sections/3_motivation.tex
\section{Background and Motivation}
\label{sec:motivation}
A floating-point tensor $\x$ is quantized into a fixed-point tensor $\xhat$ in $b$-bits using a quantization function $\xhat=q(\x;\Theta)$, where $\Theta$ denotes the quantizer parameters.
In the case of uniform affine symmetric quantization~\cite{whitepaper}, the quantization function is defined as:
\begin{equation}
\label{eq:quant}
\q(\x;\Theta) = s \left[ \text{clamp}\left(\left\lfloor \frac{\x}{\s} \right\rceil, -2^{b-1}, 2^{b-1}-1)\right) \right],
\end{equation}
where the quantizer parameters include the scaling factor $\s$, and $\lfloor \cdot \rceil$ is a rounding operation.

A floating-point convolutional function $\z= \w \ast \x$ is performed in fixed-point $\zhat=\what \ast \xhat$ by quantizing its weight and input tensors, $\w$ and $\x$, using the quantization function~\cref{eq:quant}.
The rounding and clipping operations introduce some quantization error denoted as $\epsilon=\z-\zhat$.
Reasonably, the smaller the quantization error is, the better the fixed-point model performs.
\input{figures/ranges_qnoise}

The motivation behind this work is that the inherent redundancies in video frames can be leveraged to reduce the quantization error.
Indeed, we observe that residuals exhibit a significantly lower variance than the frame activations, and can be quantized with lower error. 
\cref{fig:ranges_and_quantization_noise} supports this intuition by reporting the variances and quantization errors for frame vs. their residuals in several layers within an HRNet~\cite{hrnet} for human pose estimation:
as the figure illustrates, the aforementioned hypothesis is supported for several randomly sampled layers in the architecture.
Indeed, in the case of the normally distributed activations, it is even possible to derive formally that highly correlated samples (neighboring frames) yield residuals with lower variance:
we refer the interested reader to~\cite{difference_of_normals} for a full derivation.

Next, we further elaborate on the relation between activation variances and the quantization error, by analyzing the cases of weight and activation quantization separately.
\paragraph{Impact on weight quantization.} 
We consider the output of a linear layer for which only weights are quantized, as $\zhat = \x \ast \what$. The quantization error $\epsilon_{\w}$ can be expressed as:
\begin{align}
\epsilon_{\w} &= \x \ast \w - \x \ast \what = \x \ast (\w - \what).
\end{align}
As the equation shows, the error $\epsilon_{\w}$ is contributed to by the quantization error on weights ($\w - \what$), that is not data dependent, and by the magnitude on the input $\x$.
Considering the lower variance and average magnitude of residuals, convolving the quantized weight on them is likely to incur in a lower quantization error.
\paragraph{Impact on activation quantization.} 
\input{figures/motivation_distrib}
Similarly, for a linear layer with activation quantization, $\zhat = \xhat \ast \w$, the error can be expressed as $\epsilon_{\x} = (\x-\xhat) \ast \w = \Delta\x \ast \w$. 
Assuming no clipping, the quantization error becomes
\begin{equation}
    \Delta\x = \x - \q(\x; \Theta) = x - s \left\lfloor \frac{\x}{\s} \right\rceil = s \left( \frac{\x}{s} - \left\lfloor \frac{\x}{\s} \right\rceil \right),
\end{equation}
and is therefore bound by $-\frac{s}{2} \leq \Delta\x \leq \frac{s}{2}$. As the quantization scale $s$ is proportional to the magnitude and variance of the input $\x$, quantizing residuals with a smaller variance will lead to smaller quantization errors $\Delta x$ and $\epsilon_{\x}$.
An intuitive visualization of the relation between scale factor and quantization error is provided in \cref{fig:motivation_dist}.

%% file: figures/ranges_qnoise.tex
\begin{figure}
\centering
\resizebox{0.85\columnwidth}{!}{
\begin{tabular}{c}
\includegraphics[width=\columnwidth]{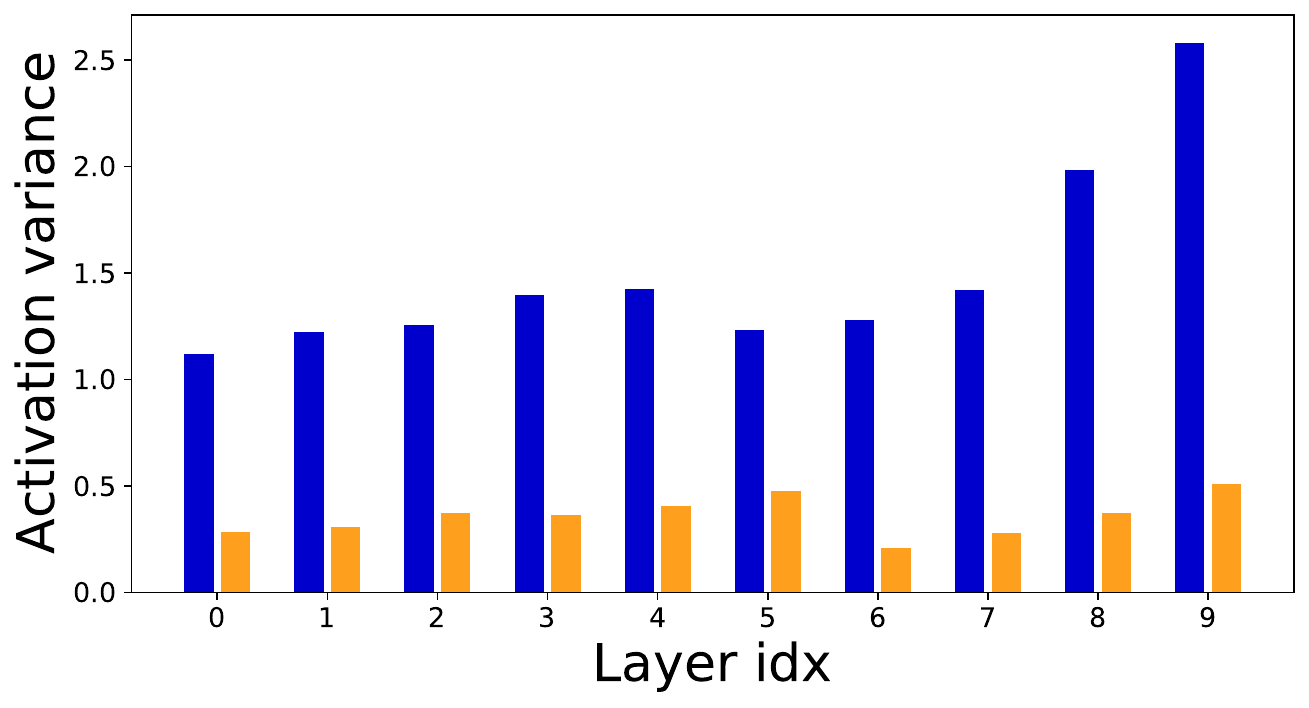}\\
\includegraphics[width=\columnwidth]{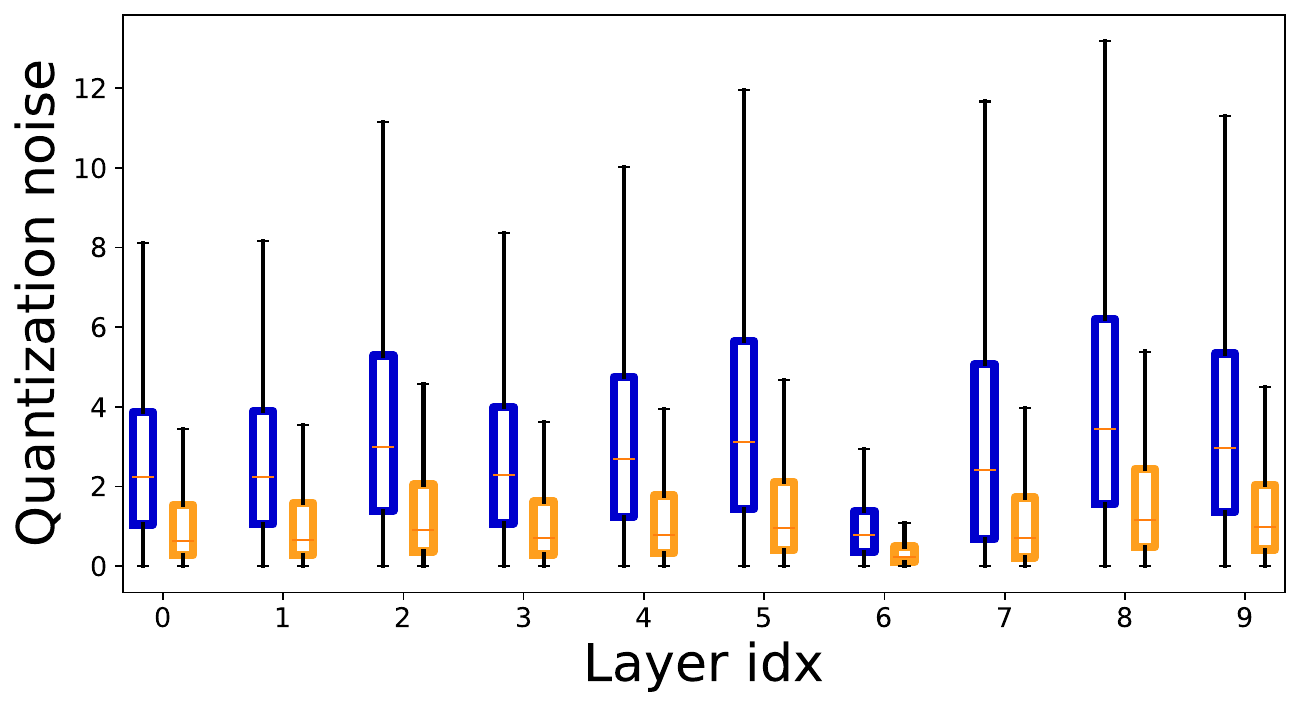}
\end{tabular}}
\caption{
\textbf{Residuals vs. Frames} for $10$ random layers from an HRNet. Residuals have a smaller variance than frames (top), translating to a lower quantization error (bottom).
}
\label{fig:ranges_and_quantization_noise}
\end{figure}

%% file: figures/motivation_distrib.tex
\begin{figure}[b]
\centering
\includegraphics[width=0.95\columnwidth]{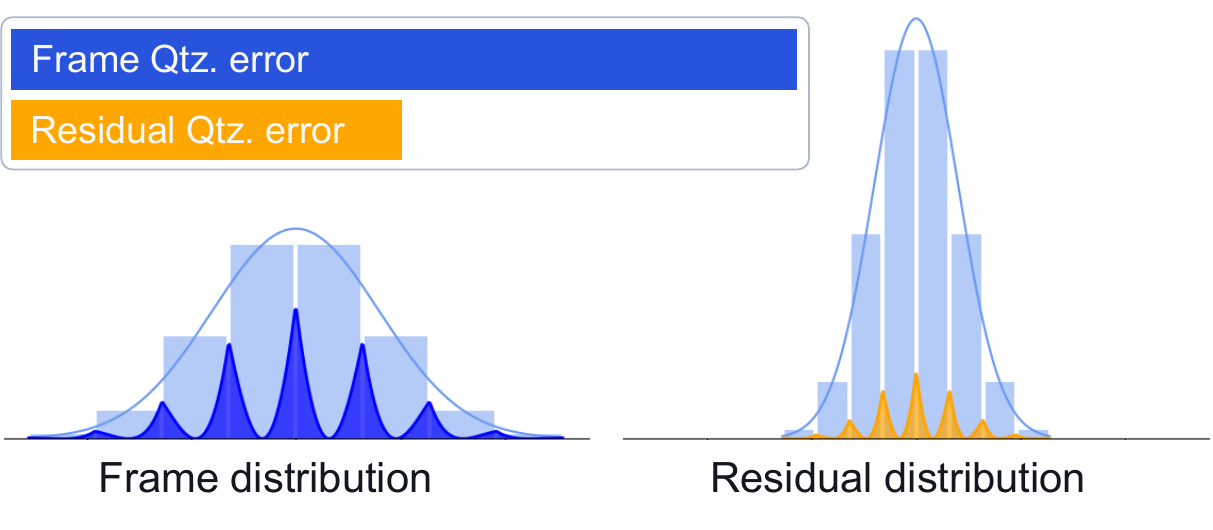}
\caption{\textbf{Impact of residual distribution on the quantization error:} The lower variance of residuals allow for smaller scale factors in their quantization (equivalent to the discrete bin-width in the plots). In turn, the expected error is lower.}
\label{fig:motivation_dist}
\end{figure}

%% file: sections/4_method.tex
\section{Residual Quantization}
\label{sec:delta_quantization}
Motivated by the intrinsic benefits of quantizing the residuals compared to the frames, we propose ResQ, a residual-based quantization scheme for video networks. 
We follow a sigma-delta formulation~\cite{sgima, skipconv, deltadist, deltacnn, eventnets}, a common approach to leverage the cross-frame redundancies, and elaborate how fixed-bit quantization can be integrated to accelerate video processing.
Using the distributive property of linear functions, the output of a convolution on a activation $\x^t$ can be formulated as:
\begin{align}
\nonumber
\z^t        &= (\x^t - \x^k) \ast \w + \x^k \ast \w, \\
\label{eq:sigma_delta}             
            &= \delta^t \ast \w + \z^k,
\end{align}
where $\delta^t$ is a residual \wrt a keyframe $\x^k$. Throughout this work, we set keyframes simply at fixed frame intervals, with a given period $T$.
~\cref{eq:sigma_delta} is performed in fixed-point using the quantization function~\cref{eq:quant}:
\begin{align}
\label{eq:sigma_delta_quant} 
\zhat^t     &= \q(\delta^t;\Theta_a) \ast \q(\w;\Theta_w) + \zhat^k, \\
\nonumber
\zhat^k     &= \q(\x^k;\Phi_a) \ast \q(\w;\Phi_w),
\end{align}
where $\Phi_w$ and $\Phi_a$ denote the weight and activation quantization parameters for keyframes, respectively.
The same set of quantization parameters is shared for all the keyframes. 
Similarly, $\Theta_w$ and $\Theta_a$ denote the weight and activation quantization parameters for residual frames.
These can be shared for all residual frames or can be dynamically adjusted based on the residual content as discussed in~\cref{sec:dynamic}. 
Separating the quantization parameters between keyframes and residuals allows amortizing high-precision computation on the keyframe with low-precision computations on residuals, for an overall lower quantization error.
Inference with our model is represented in \cref{fig:model_delta_quantization}.
\input{figures/model}
\subsection{Estimating Quantizer Parameters}
\label{sec:calibration}
For quantizing both keyframes and residuals, we rely on a uniform affine symmetric scheme in~\cref{eq:quant}.
The scale factor $s$ depends on the activation range of the quantizer, identified by $\rmin,\rmax$, as follows:
\begin{equation}
\label{eq:scale_factor}
\s=s(\rmax,\rmin)=\frac{2 \max(\rmax, - \rmin)}{2^b - 1}.
\end{equation}
At any convolution layer, we employ two different range setters for quantizing the input $\x$ and the weights $\w$.
Specifically, for weights, we employ a simple min-max strategy,
\begin{equation}
\rmax^{\w} = \max(\w), \quad \rmin^{\w} = \min(\w),
\end{equation}
and directly compute the quantized weights $\what$.

For estimating the range of activations $\rmin^{\x},\rmax^{\x}$, we first collect exemplar inputs to the layer $\{\x_i\}_1^c$, by feeding $c$ calibration samples to the model.
We then concatenate them into a batch $\X$, and build a line-search space $\mathcal{S}$ between its minimum and maximum, with $r$ candidate points, as $\mathcal{S}=\text{linspace}(\min(\X),\max(\X),r)$.
We then search the range in $\mathcal{S}$, by minimizing the following objective, where the scale factor $s$ is derived from the range using \cref{eq:scale_factor}:
\begin{equation}
\label{eq:range_activation}
\argmin_{\rmax^{\x},\rmin^{\x} \in \mathcal{S}} \|\X \ast \w - \q \bigl( \X;\Theta=s(\rmax^{\x},\rmin^{\x})\bigr) \ast \what \|_F.
\end{equation}
\subsection{Dynamic Residual Quantization}
\label{sec:dynamic}
\input{figures/jhmdb_delta_vs_act}
As discussed in~\cref{sec:motivation}, residuals have a lower variance compared to frames, thus are typically easier to quantize.
However, not all residuals have similar characteristics, as they depend on the video content.
For static scenes, residuals usually have a low magnitude, whereas in the presence of motion they can exhibit a higher dynamic range, making their quantization more challenging.
This motivates us to design a \textit{dynamic residual quantization} scheme that adaptively adjusts the precision to the amount of changes occurring in a video.
To this aim, instead of using one set of parameters for quantizing all timesteps, we adjust the quantizer parameters based on the residual content.
We consider this adjustment for the activation quantizer parameters only ($\Theta_a$ in~\cref{eq:sigma_delta_quant}) as weights are static and input independent.

Dynamic quantization can be performed at various granularities,~\eg at frame-, region-, or pixel-level. 
We opt for the pixel-level scheme, which shares the same quantizer along the channel dimension for every pixel.
This gives us full flexibility in adapting to any spatial differences in residual frames while still being computationally efficient.
In dynamic residual quantization, every layer has a single quantizer for weights $\Theta_w$, and a \emph{pool} of quantizers, $\{\Theta_a^1 \dots \Theta_a^n\}$, ordered from low to high bit-width, available for input residuals.
All quantizers are fit separately in the calibration stage, as explained in \cref{sec:calibration}, identically to the case of static residual quantization.

During inference, given a residual $\dlt \in \mathbf{R}^{c\times h \times w}$, we produce an index map $\Pi\in \{1 \dots n\}^{h \times w}$ as $\Pi = \pi(\dlt)$,
where $\pi$ is a policy function selecting the optimal quantizer per pixel. 
We rely on the quantization error as the criteria to design the policy function.
Specifically, for each quantizer $\Theta^i_a$ from the pool, we compute an error map $\epsilon_i \in \mathbf{R}^{h \times w}$ as:
\begin{equation}
\label{eq:policy_oracle}
\epsilon_i = \| \dlt \ast \what - \q(\dlt;\Theta_a^i) \ast \what\| = \|(\dlt - \q(\dlt;\Theta_a^i)) \ast \what\|,
\end{equation}
where the norm is to be intended pixel-wise.
However, the estimation in \cref{eq:policy_oracle} is troublesome, as it requires projecting the quantization error on residuals, $\dlt - q(\dlt;\Theta_a^i)$ through $\what$.
Using Young's convolution inequality~\cite{young}, we approximate the error map by multiplying the norms of $\dlt - q(\dlt;\Theta_a^i)$ and $\what$, where the former has to be intended pixel-wise:
\begin{equation}
\label{eq:policy_approximate}
\epsilon_i \leq \| \dlt - \q(\dlt;\Theta_a^i) \| \| \what \|.
\end{equation}
The approximate error in~\cref{eq:policy_approximate} serves as an upper bound of the objective in \cref{eq:policy_oracle}~\cite{young}.
Once an error map $\epsilon_i$ is approximated for every quantizer in the pool $\Theta_a^1 \dots \Theta_a^n$, the policy function starts from the lowest precision and iteratively considers the reduction in quantization error given by selecting the next (higher bits) quantizer:
\begin{equation}
\label{eq:policy}
\Pi = \min_{i=1,\dots,n} i \quad \text{s.t.} \quad \epsilon_i - \epsilon_{i+1} < \tau.
\end{equation}
Intuitively, for every residual pixel we consider the reduction in quantization error granted by the next precision $\Theta_a^{i+1}$ and we stop as soon as the gap is smaller than a predetermined threshold $\tau$.
For a visualization of the decisions taken by the policy function, we refer the reader to \cref{fig:policy decisions}.

%% file: figures/model.tex
\begin{figure}[t]
\centering
\includegraphics[width=0.9\columnwidth]{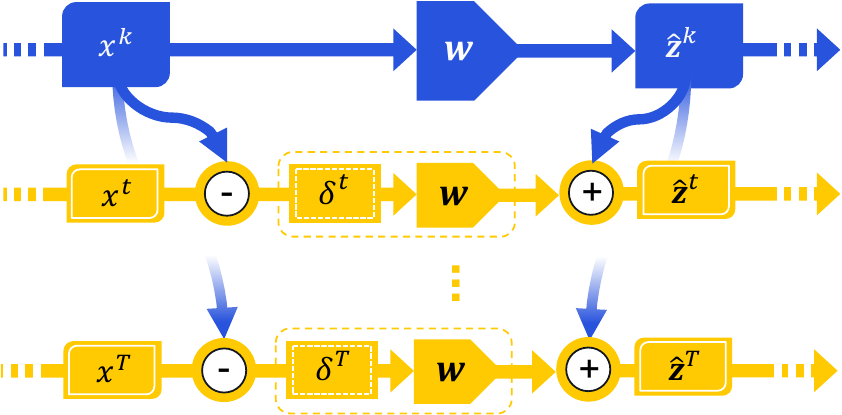}
\caption{\textbf{\method at inference.} The key-frame activation and weights, in blue, are quantized at higher precision, whereas the residual activations and weights highlighted in orange are quantized at lower precision. The activations at residual timesteps are obtained by summation with the key-frame activations. This scheme is applied for all layers throughout the model.}
\label{fig:model_delta_quantization}
\end{figure}

%% file: figures/jhmdb_delta_vs_act.tex
\begin{figure*}[t]
\centering
\includegraphics[width=0.75\columnwidth]{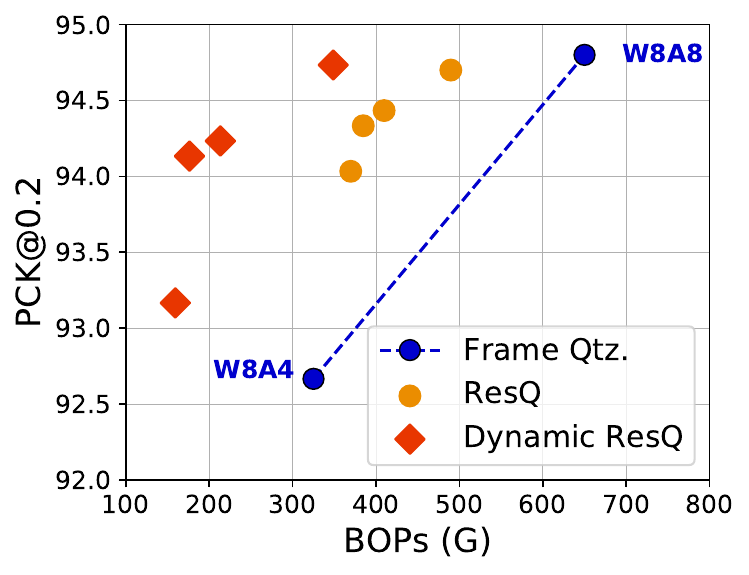}
\includegraphics[width=0.75\columnwidth]{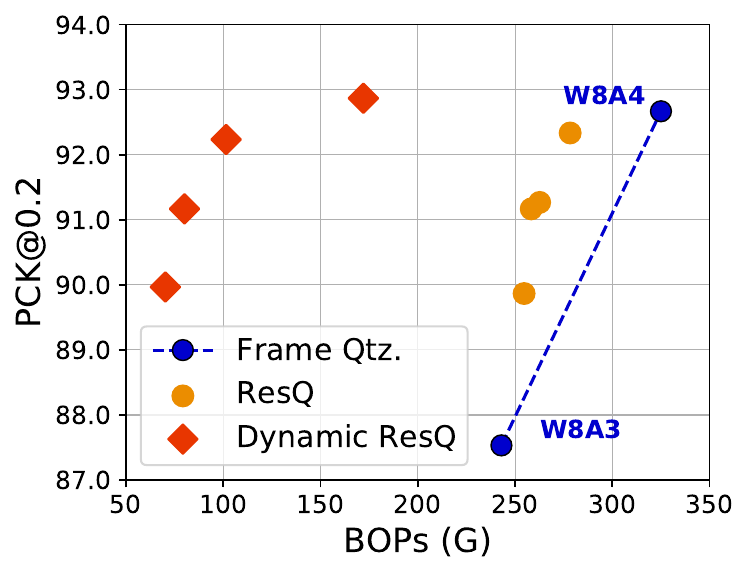}
\caption{\textbf{Comparison of quantization schemes on JHMDB}. Curves generated by sweeping $T=2,4,6,8$.}
\label{fig:jhmdb_delta_vs_act}
\end{figure*}

%% file: sections/5_experiments.tex
\section{Experiments}
\label{sec:exps}
\paragraph{Efficiency metric and evaluation.}
As current deep learning frameworks (\eg PyTorch) currently support 32, 16, and 8-bit kernels only, we follow the common practice in quantization~\cite{bops,bayesianbits} and rely on Bit Operations (BOPs) to measure the computational costs:
BOPs are computed by re-weighting the multiply-accumulate (MAC) count of a model accounting for bit-widths of weights and activations.
Since residual quantization utilizes different bit-width for the keyframe and the residual frames, we report precision in a unified notation: for instance, \amprec{W8A8}{W8A4} indicates bit-width of 8-8, for weights-activation of keyframes, and bit-width of 8-4 for weight-activation of residual frames.

Following~\cite{hu2020tdnet,deltacnn,skipconv}, we split each video into sequences of length $T$, whose first frame is treated as keyframes. 
BOPs and accuracy metrics are averaged over all frames in a sequence.
For the \dymethod~experiments, the policy function is considered in the BOP count.
\paragraph{Quantization methods.}
We conduct our experiments with both PTQ and QAT, as explained in \cref{sec:related}.
For PTQ, we use $c=64$ calibration samples and $r=20$ line search points for selecting the activation quantization range.
For QAT, we initialize the scale factor $s$ of the quantizer via PTQ, and propagate gradients through the rounding operation in \cref{eq:quant} by means of a straight-through estimator~\cite{straightthrough}. 
We either apply the same scaling factor $s$ for the whole tensor, \ie tensor quantization, or have separate $s$ for each channel of the weight tensor, \ie channel quantization.
Unless specified otherwise, we utilize tensor quantization.
For the dynamic policy, we fix $\tau=0.0003$ for all experiments.
\input{sections/5_1_experiments_pose}
\input{sections/5_2_experiments_segmentation}
\input{sections/5_3_experiments_vos}

%% file: sections/5_1_experiments_pose.tex
\subsection{Human Pose Estimation}
\label{sec:exp_pose}
\input{tables/jhmdb_sota}
\input{figures/dynamic_policy_decisions}
\paragraph{Dataset and accuracy metric.}
We use JHMDB~\cite{jhmdb}, consisting of a collection of 316 clips, comprising more than 11,000 frames.
Every frame is labeled with the location of 15 body joints, for a single person in the scene.
The dataset comes with three different train/test splits, that we use accordingly with the standard protocol: every reported operating point is an average over all splits.
As a performance metric, we rely on the standard PCK metric, with threshold $\alpha=0.2$, as utilized in~\cite{song2017thin,luo2018lstm,nie2019dynamic,skipconv}.

\paragraph{Implementation details.}
As a starting floating point model, we rely on HRNet-w32~\cite{hrnet_pose}, pretrained on MPII Human Pose~\cite{mpi}, that we fine-tune on every split for 100 epochs using the Adam optimizer~\cite{adam} with a batch size of 16 images.
The learning rate is initialized to $1e-3$ and decayed by a factor 10 at epochs 40 and 80.
We follow standard training augmentations such as scaling, rotations, flipping and cropping, following the implementation of~\cite{xiao2018simple}.
\paragraph{\method vs. frame quantization.}
We evaluate the effectiveness of \method~by comparing it with Frame Quantization, the standard quantization strategy processing frames independently.
We quantize all models using PTQ with the same setting, in terms of range setters and calibration samples, to foster a fair comparison.
\cref{fig:jhmdb_delta_vs_act} reports the results for \amprec{W8A8}{W8A4} (left) and \amprec{W8A4}{W8A3} (right) bit-width configurations. 
\method~achieves a better trade-off between accuracy vs. efficiency with respect to Frame Quantization, supporting our core hypothesis that quantizing the residuals leads to a smaller quantization error.
\paragraph{Dynamic vs. static \method.}
We evaluate the effectiveness of the dynamic residual quantization, proposed in~\cref{sec:dynamic}, by comparing \dymethod~vs. \method in configurations of \amprec{W8A8}{W8A\{0,4,8\}} as reported in \cref{fig:jhmdb_delta_vs_act} (left).
More specifically, we use 8-bit weights while dynamically adjusting the activation bit-width among 0, 4, and 8 bits.
Similarly, \cref{fig:jhmdb_delta_vs_act} (right) reports the performance for \amprec{W8A8}{W8A\{0,3,4\}}.
As reported, \dymethod outperforms both \basemethod~and ResQ, advocating for the benefits of dynamically selecting the bit-width given the residual content.
To gain more insights about the policy function, we represent in color coding its decisions in \cref{fig:policy decisions}.
As the figure shows, moving objects are quantized to a higher bit-width, whereas background and static areas tend to be assigned low-precision, contributing to a reduction in BOP cost without sacrificing accuracy.
\input{tables/cityscapes_sota}
\input{figures/cityscapes/ddrnet23s_ptq}
\input{figures/cityscapes/cityscapes_qualitative}
\paragraph{Comparison with the state-of-the-art.}
\cref{tab:jhmdb} compares our method with the current state-of-the-art for efficient human pose estimation in video: weight compression~\cite{spatial_svd,weight_svd} and pruning~\cite{louizos2018learning} methods, as well as dynamic kernel distillation (DKD)~\cite{nie2019dynamic} and Skip-Convolution~\cite{skipconv} that leverage temporal redundancies to accelerate video inference. Except for DKD, all other methods use the same HRNet-w32~\cite{hrnet_pose} backbone.
The BOPs of competing methods have been computed from the MACs reported in the corresponding papers and assuming lossless quantization to W8A8.

The results confirm that \dymethod achieves the lowest computational cost, $176$ GBOPs, whilst maintaining a high accuracy of $94.1$. 
Such results are only $1\%$ less accurate than the highest PCK reported by Skip-Conv while being $\sim 3\times$ more efficient.
Moreover, both \method and \dymethod perform favorably compared to the alternatives in terms of accuracy vs. efficiency trade-off.
Finally, it is worth noting that in this experiment we rely only on PTQ and do not use any form of fine-tuning, contrary to all competing methods that need a supervised training procedure.

%% file: tables/jhmdb_sota.tex
\begin{table}[b]
\centering
\resizebox{\columnwidth}{!}{
\begin{tabular}{lccccccccc}
\toprule
& \textbf{GBOPs} & \textbf{Hea}. & \textbf{Sho}. & \textbf{Elb}. & \textbf{Wri}. & \textbf{Hip} & \textbf{Kne}. & \textbf{Ank}. & \textbf{Avg}\\
\midrule
DKD\etal~\cite{nie2019dynamic} & 553 & 98.3 & 96.6 & 90.4 & 87.1 & 99.1 & 96.0 & 92.9 & 94.0\\
S-SVD~\cite{spatial_svd} & 322 & 97.9 & 96.9 & 90.6 & 87.3 & 98.7 & 95.3 & 91.1 & 94.3\\
W-SVD~\cite{weight_svd} & 325 & 97.9 & 96.3 & 87.2 & 82.8 & 98.1 & 93.2 & 88.8 & 92.4\\
L0~\cite{louizos2018learning} & 208 & 97.1 & 95.5 & 86.5 & 81.7 & 98.5 & 92.9 & 88.6 & 92.1\\
Skip-Conv\cite{skipconv} & 539 & 98.7 & 97.7 & \textbf92.0 & 88.1 & 99.3 & 96.6 & 91.0 & \textbf{95.1}\\
\midrule
\textbf{\method} & 384 & 98.3 & 97.2 & 90.9 & 87.0 & 99.0 & 95.1 & 91.0 & 94.3\\
\textbf{\dymethod} & \textbf{176} & 98.2 & 97.0 & 91.0 & 87.1 & 98.7 &  94.6 & 90.5 & 94.1\\
\bottomrule
\end{tabular}}
\caption{\textbf{Comparison to state-of-the-art on JHMDB.} \dymethod achieves the best accuracy vs. efficiency trade-off.}
\label{tab:jhmdb}
\end{table}

%% file: figures/dynamic_policy_decisions.tex
\bgroup
\setlength{\tabcolsep}{0.5pt}
\renewcommand{\arraystretch}{0.2}
\begin{figure*}[tbh]
\centering
\resizebox{0.95\textwidth}{!}{
\begin{tabular}{ccccccc}
$\Delta$t=1 & $\Delta$t=2 & $\Delta$t=3 & $\Delta$t=4 & $\Delta$t=5 & $\Delta$t=6 & $\Delta$t=7\\
\includegraphics[width=0.15\textwidth]{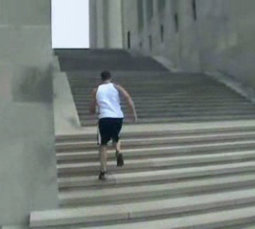}&
\includegraphics[width=0.15\textwidth]{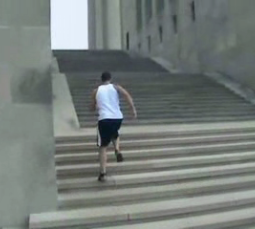}&
\includegraphics[width=0.15\textwidth]{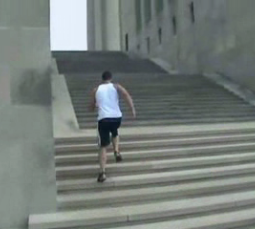}&
\includegraphics[width=0.15\textwidth]{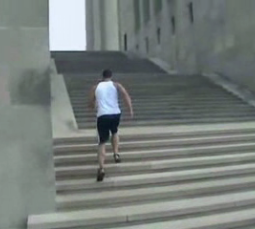}&
\includegraphics[width=0.15\textwidth]{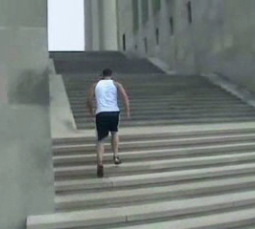}&
\includegraphics[width=0.15\textwidth]{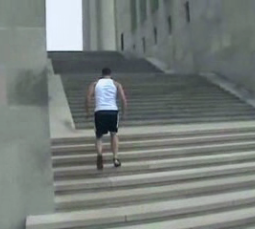}&
\includegraphics[width=0.15\textwidth]{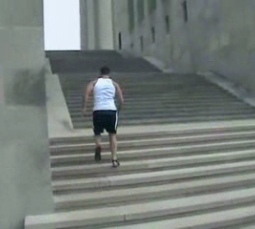}\\
\includegraphics[width=0.15\textwidth]{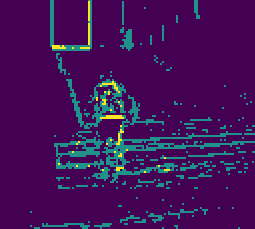}&
\includegraphics[width=0.15\textwidth]{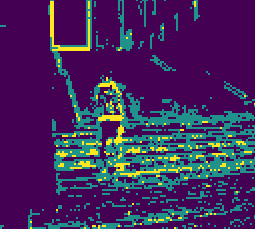}&
\includegraphics[width=0.15\textwidth]{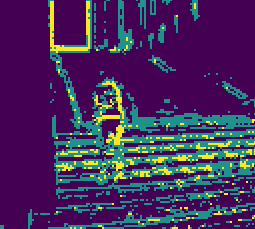}&
\includegraphics[width=0.15\textwidth]{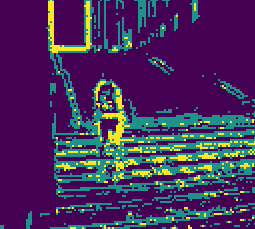}&
\includegraphics[width=0.15\textwidth]{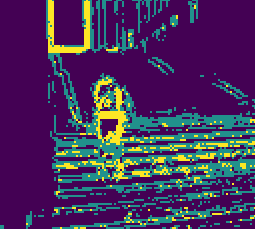}&
\includegraphics[width=0.15\textwidth]{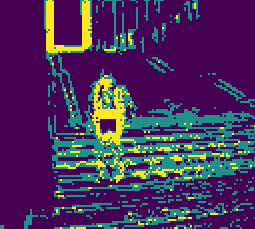}&
\includegraphics[width=0.15\textwidth]{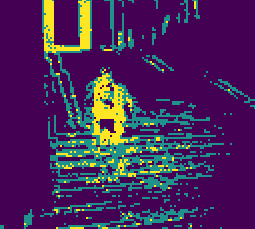}
\end{tabular}}
\caption{\textbf{\dymethod policy.} We represent policy decisions for \texttt{conv2} in HRNet. Dark blue, light blue and yellow pixels color code pixels that are quantized to 0, 4, and 8 bits respectively. Few regions require high precision, and many stationary regions are processed in low precision. Selected precision generally increases with the distance to the keyframe ($\Delta t$).}
\label{fig:policy decisions}
\end{figure*}
\egroup

%% file: tables/cityscapes_sota.tex
\begin{table}[b]
\begin{center}
\resizebox{0.99\columnwidth}{!}{
\begin{tabular}{lllcc}
\toprule
\textbf{Model} & \textbf{Backbone} & \textbf{Bit-width} &    \textbf{BOPs (T)} $\downarrow$  &  \textbf{mIoU} $\uparrow$ \\
\midrule
TDNet~\cite{hu2020tdnet} & PSPNet~\cite{zhao2017pspnet} & W8A8* & 34.5 & \textbf{79.9} \\
TDNet~\cite{hu2020tdnet} & BiseNet~\cite{bisenetv2} & W8A8* & 6.4 & 76.4 \\
DFF~\cite{zhu17dff} & ResNet-101~\cite{resnet} & W8A8*                 & 7 & 69.2\\
Delta Dist.~\cite{deltadist} & HRNet-w18s~\cite{hrnet}  & W8A8*     & 2.2 & 75.7 \tinyspace\\
\textbf{\method} & HRNet-w18s & \amprec{W4A4}{W3A3}     & \textbf{0.9} & 76.9\tinyspace\\
\midrule
Delta Dist. ~\cite{deltadist} & DDRNet-39~\cite{ddrnet}  & W8A8*     & 9.0 & 79.9\tinyspace\\
\textbf{\method} & DDRNet-39 & \amprec{W8A8}{W4A4}     & 9.0 & \textbf{80.8}\tinyspace\\
\textbf{\method} & DDRNet-39 & \amprec{W4A4}{W3A3}    & \textbf{3.2} & 80.0\tinyspace\\
\midrule
Delta Dist.~\cite{deltadist} & DDRNet-23~\cite{ddrnet}  & W8A8*     &4.6 & 78.9 \tinyspace\\
\textbf{\method} & DDRNet-23 &  \amprec{W8A8}{W4A4}     & ~4.6 & ~\textbf{79.3} \tinyspace\\
\textbf{\method} & DDRNet-23 &  \amprec{W4A4}{W3A3}     & \textbf{1.6} & 78.8  \tinyspace\\
\midrule
Skip-Conv~\cite{skipconv} & DDRNet-23s~\cite{ddrnet} & W8A8*          & 1.7 & 75.5  \tinyspace\\
Delta Dist. ~\cite{deltadist}& DDRNet-23s & W8A8*     & 1.1 & 76.2 \\
\textbf{\method} & DDRNet-23s &  \amprec{W8A8}{W4A4}     & 1.1 & \textbf{77.3} \\
\textbf{\method} & DDRNet-23s &  \amprec{W4A4}{W3A3}     & \textbf{0.4} & ~76.2 \\
\midrule
\end{tabular}}
\end{center}
\caption{\textbf{Comparison with state-of-the-art on Cityscapes.} \method outperforms alternatives with similar or less BOPs. *Models are assumed lossless in performance when quantized to 8 bits.
}
\label{tab:sota_segmentation} 
\end{table}

%% file: figures/cityscapes/ddrnet23s_ptq.tex
\begin{figure*}[t]
\centering
\includegraphics[width=0.85\columnwidth]{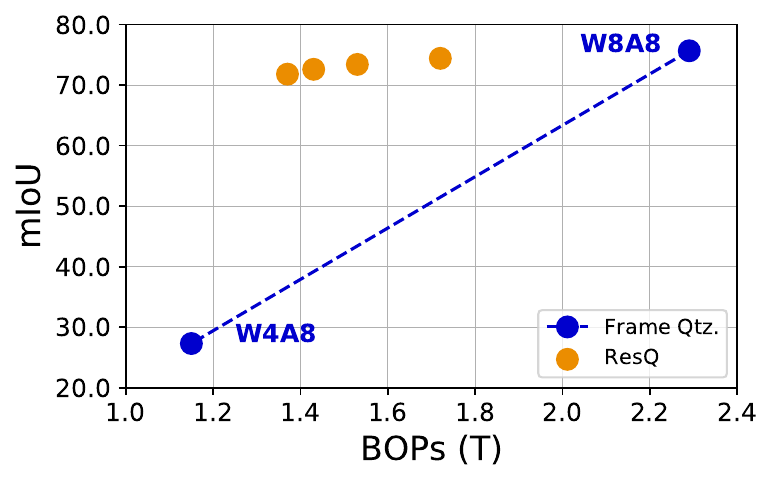}
\includegraphics[width=0.85\columnwidth]{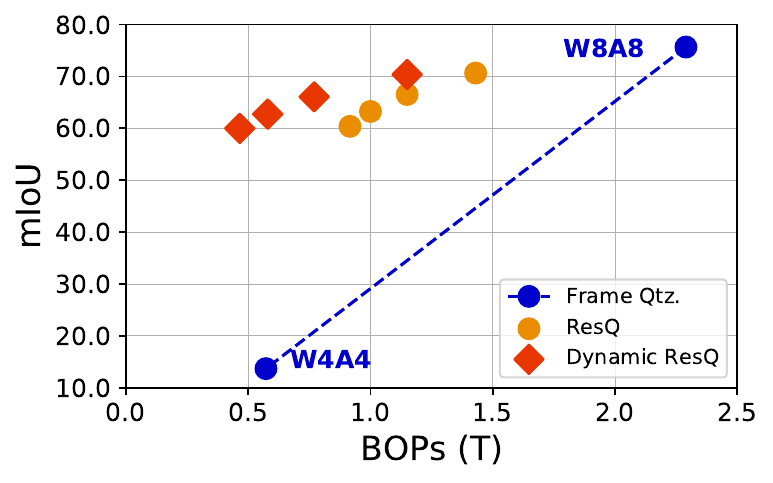}
\caption{
\textbf{Frame vs. Residual Quantization on Cityscapes.} The base model is DDRNet-23 slim. Points are generated for $T=2,3,4,5$. 
}
\label{fig:cityscapes_ddrnet23s_ptq}
\end{figure*}

%% file: figures/cityscapes/cityscapes_qualitative.tex
\bgroup
\setlength{\tabcolsep}{1pt}
\begin{figure*}[t]
\centering
\includegraphics[width=0.95\textwidth]{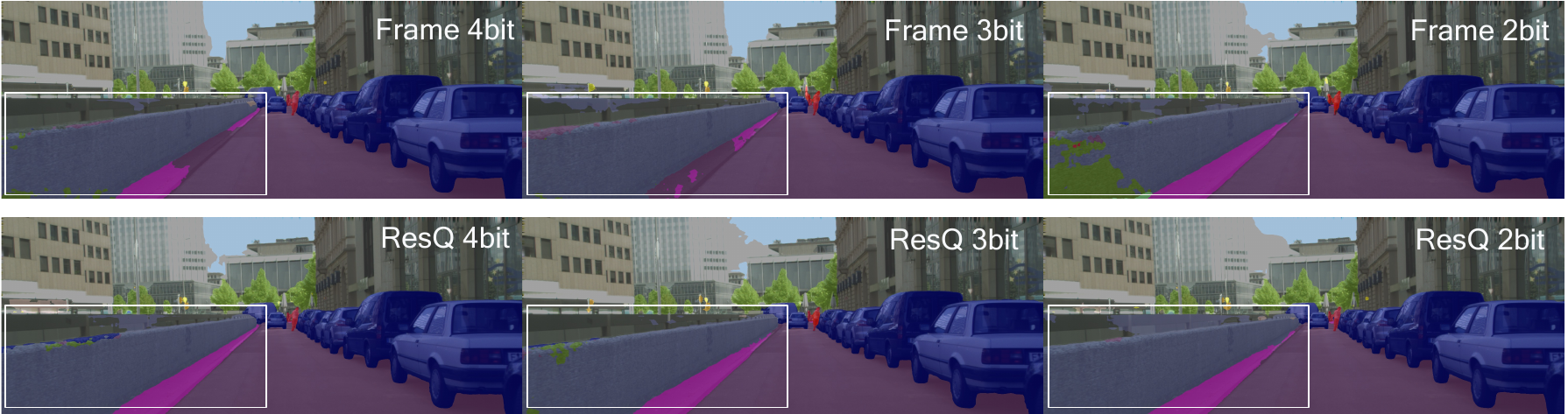}
\caption{\textbf{Qualitative comparison} between \method and Frame Quantization. \method takes advantage of information from the previous keyframe to improve the segmentation of the scene.}
\label{fig:cityscapes_qualitative}
\end{figure*}
\egroup

%% file: sections/5_2_experiments_segmentation.tex
\subsection{Semantic Segmentation}
\label{sec:exp_semseg}
\input{tables/davis_frame_vs_residual}
\paragraph{Dataset and metric.}
We use the Cityscapes dataset~\cite{cityscapes} and rely on the standard training and validation splits with 2,975 and 500 clips, respectively. 
Cityscapes provides pixel-level annotations into 19 classes for one frame per snippet.
Following~\cite{deltadist}, we extract the per-frame pseudo-annotations required to train video models by applying an off-the-shelf segmentation network~\cite{tao2020hierarchical} on unannotated frames in the training set.
We evaluate the accuracy using mean Intersection-over-Union (mIoU) and perform inference by selecting the keyframe at intervals of $T=3$ frames. 
To deal with sparse labels in the validation set, we follow the protocol in~\cite{hu2020tdnet,deltadist} and average evaluations for all clips in which each labeled frame falls in position $t\leq T$.

\paragraph{Implementation details.}
We conduct our experiments on DDRNet~\cite{ddrnet} and HRNet-w18-small~\cite{hrnet} backbones, and perform experiments with both PTQ and QAT quantizers.
When using QAT, we fine-tune the model quantized with PTQ using the original training protocol~\cite{ddrnet, hrnet}, except for the learning rate, which is lowered to $5e-4$.
As such, we train with 4 GPUs, a batchsize of 12 using SyncBN for 484 epochs, and we employ online hard example mining~\cite{shrivastava2016training}.

\paragraph{Comparison of quantization schemes}
\cref{fig:cityscapes_ddrnet23s_ptq} compares \basemethod and \method on DDRNet-23 slim, for two different bit-width configurations (\amprec{W8A8}{W4A8} and \amprec{W8A8}{W4A4}).
Consistent with the previous results on human pose estimation, we observe that \method outperforms Frame Quantization.
Furthermore \dymethod (\amprec{W8A8}{W4A\{0,4,8\}}) outperforms \method and largely recovers the catastrophic degradation hampering \basemethod in low bit-width regimes (W4A4).

\cref{fig:cityscapes_qualitative} qualitatively compares \basemethod and \method at 4, 3, and 2 bits.
As seen on the highlighted regions, our proposal takes advantage of the high precision representation for the keyframe to better segment the scene.
\paragraph{Comparison with the state-of-the-art}
We compare \method with prior solutions that are devised to accelerate video inference:
TDNet~\cite{hu2020tdnet}, DFF~\cite{zhu17dff}, Skip Convolutions~\cite{skipconv} and Delta Distillation~\cite{deltadist}.
Again, we assume lossless quantization to 8 bits for competing methods and compare against their upper bound accuracies. \method relies on QAT and channel quantization.
As shown in \cref{tab:sota_segmentation}, our proposal achieves a better accuracy vs. efficiency trade-off against existing models.
More specifically, \method improves over Delta Distillation by 0.9, 0.4, and 1.1 points in mIoU for DDRNet-39, DDRNet-23, DDRNet-23s, respectively.
Interestingly, we notice our model improves in mIoU over the floating point model we are optimizing.
We ascribe this behavior to the fact that, by propagating keyframe representations to future timesteps, with \method the model successfully learns to leverage the temporal context.

\paragraph{Temporal stability.}
\input{figures/cityscapes/cityscapes_temporal_robustness}
Temporal degradation can be a concern in keyframe based models. 
To assess this aspect, we consider \method on DDRNet23s on Cityscapes, with QAT precision \amprec{W8A8}{W4A4}.
\cref{fig:temporal_robustness} plots its segmentation mIoU against the distance to the last keyframe, along with results from other video efficiency models~\cite{zhu17dff,skipconv,deltadist}. 
As shown, our model proves more robust than competing methods, with improvements ranging from 0.5 to 1 point. 

%% file: tables/davis_frame_vs_residual.tex
\begin{table*}[t]
\centering
\resizebox{0.9\textwidth}{!}{
\begin{tabular}{llcccccc}
\toprule
\multirow{2}{*}{\textbf{Model}} & \multirow{2}{*}{\textbf{Bit-width}}&\multicolumn{3}{c}{\textbf{Single Object VOS} (DAVIS-2016)} & \multicolumn{3}{c}{\textbf{Multi Object VOS} (DAVIS-2017)}\\
 && \textbf{$\mathcal{J}$-Mean} $\uparrow$ & \textbf{$\mathcal{F}$-Mean} $\uparrow$ & \textbf{BOPs (T)} $\downarrow$& \textbf{$\mathcal{J}$-Mean} $\uparrow$ & \textbf{$\mathcal{F}$-Mean} $\uparrow$ & \textbf{BOPs (T)} $\downarrow$\\
\midrule
\multirow{3}{*}{STM~\cite{stm}} & W8A8 & 83.8 & 85.9 & 10.07 & 75.7 & 81.1 & 19.09 \\
& W4A8 & 00.1 & 00.5 & 5.04 & 01.8 & 03.5 & 9.54 \\
& W8A4 & 00.9 & 06.0 & 5.04 & 03.0 & 11.6 & 9.54 \\
\midrule
\multirow{2}{*}{+\textbf{\method}}& \amprec{W8A8}{W4A8} & 73.4 & 76.0 & 6.74 & 67.3 & 72.8 & 12.78\\ 
& \amprec{W8A8}{W8A4} & 70.2 & 71.9 & 6.74 & 65.3 & 70.5 & 12.78\\
\midrule
+\textbf{\dymethod} & \amprec{W8A8}{W8A[0,4,8]} & 82.8 & 83.8 & 6.51 & 75.1 & 79.3 & 11.33\\
\bottomrule
\end{tabular}}
\caption{\textbf{VOS results} on the DAVIS 2016 and 2017 benchmarks. The different BOP counts for the two splits can ascribed to the multi-object nature of DAVIS-2017, requiring multiple joint encodings of frames and masks~\cite{stm}. Low-bit quantization disrupts the baseline, while \method and \dymethod retain segmentation capabilities at lower precision.}
\label{tab:davis}
\end{table*}

%% file: figures/cityscapes/cityscapes_temporal_robustness.tex
\begin{figure}[t]
\centering
\includegraphics[width=0.85\columnwidth]{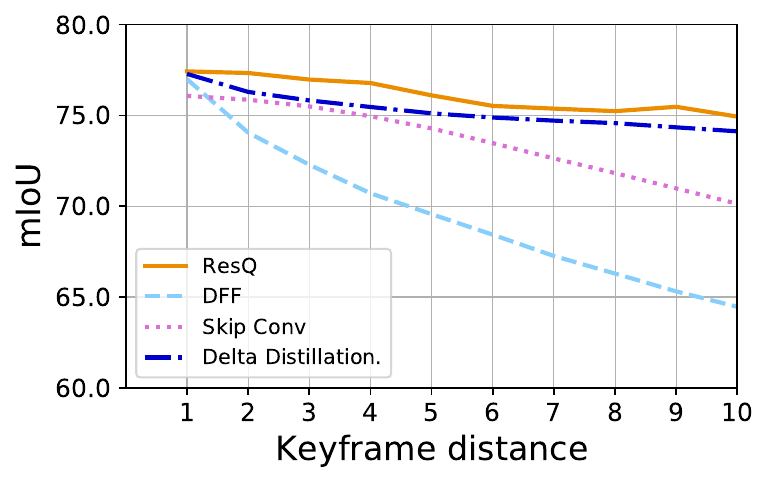}
\caption{\textbf{Temporal stability}. We measure the impact of using residuals for up to 10 frames and compare against similar video efficiency models based on keyframe and residual processing.}
\label{fig:temporal_robustness}
\end{figure}

%% file: sections/5_3_experiments_vos.tex
\subsection{Video Object Segmentation}
\label{sec:exp_vos}
\paragraph{Dataset, base model and metric.}
We conduct additional experiments with Video Object Segmentation (VOS) on the DAVIS dataset.
More specifically, we experiment with semi-supervised VOS, where  objects are segmented on the first frame of a clip and their masks are propagated to successive frames.
As a base model, we choose Space-Time Memory Networks (STM)~\cite{stm}.
We use the released weights, obtained by the authors by training on Youtube-VOS~\cite{youtubevos}, and we quantize the encoder and decoder modules with PTQ, whilst leaving the memory operations at full precision.
We test our model on the validation sets of DAVIS-2016~\cite{davis16}, composed of 20 clips each with a single target object, as well as DAVIS-2017~\cite{davis17}, where multiple objects are targeted in 30 videos.
We rely on the official evaluation SDK and report performance in terms of region ($\mathcal{J}$) and boundary ($\mathcal{F}$) accuracy (as defined in~\cite{davis16}).
\paragraph{Results.}
Performance of both \basemethod and \method ($T=3$) on the 2016 and 2017 splits are reported in \cref{tab:davis}.
We found the model hard to quantize, possibly due to quantization noise interfering with its memory operations.
Indeed, quantization to 8 bits is already lossy w.r.t.~the floating point model (4.9 and 3.5 drop in $\mathcal{J}$-Mean for 16 and 17 benchmarks respectively).
As the Table shows, when quantizing to 4 bits, \basemethod demeans the model, whereas \method and \dymethod retain performance.
Remarkably, \dymethod suffers minor drops in $\mathcal{J}$-Mean of 1 point and 0.6 points for the single and multi-object benchmark respectively, allowing to save approximately 35 to 40\% BOPs.

%% file: sections/6_conclusions.tex
\section{Conclusion}
We motivated - and supported with theoretical insights - that whenever running quantized models on video streams a lower quantization error can be obtained by using residuals of frame representations.
We have therefore proposed \method, a technique for amortizing expensive high precision computations on a keyframe over several residual frames, by processing the latters with low precision.
We further introduced a dynamic extension, where precision for each residual pixel is adjusted on the fly, based on content of the residual itself.
Through several experiments, we have empirically validated that \method can obtain favorable results in terms of accuracy and efficiency, outperforming existing methods in this trade-off in three different perception tasks.
\paragraph{Limitations.}
We highlight some potential limitations of our model to be tackled by future work.
One downside to \method is that it requires the propagation of representations to future timesteps, leading to a memory overhead potentially impacting latency in memory-bound applications.
Moreover, implementing location-specific quantized operations is not trivial and requires specialized hardware or gather-scatter implementations of convolutions, similar to what previously used for sparse processing in~\cite{sbnet,verelst2020dynamic}.
Finally, we note how \method is able to reduce the amortized cost of video processing, yet the peak BOPs is not reduced. 

%% file: sections/supp_text/additional_exps.tex
\section{Additional experiments}
\input{figures/cityscapes/combined_qat.tex}
\paragraph{Semantic segmentation: QAT.}
When labeled data is available, Quantization Aware Training (QAT) is a viable option to recover performance lost during Post Training Quantization (PTQ).
We hereby report additional results for QAT of segmentation models on Cityscapes.
More specifically, we consider the backbones DDRNet23 slim, DDRNet23, DDRNet39 and HRNet-w18-small and employ, for each of them, four different precisions (8, 4, 3 and 2 bits).
Moreover, we implement several \method models alternating between such precisions in keyframes and residual frames, in cycles of $T=3$ timesteps.
Results for this experiment are reported in \cref{fig:cityscapes_qat}, where we represent in blue the baseline results of frame quantization, and in other colors the result from \method models.
In this respect, the colors representing \method results identify the precisions it operates at (w.r.t. the dashed lines between blue points): for example, the orange points represent \method models with precision \amprec{W8A8}{W4A4}.
As the figure shows, even in the presence of QAT our proposal can outperform the baseline of frame quantization.
Interestingly, in many cases our final amortized IoU even exceeds the one of the keyframe: we ascribe this gap to the fact that \method is a video model, and can learn temporal dynamics during the QAT fine-tuning stage.

To further compare the benefit that fine-tuning can bring when using residual quantization, we directly compare \method results in the case of PTQ and QAT in \cref{tab:ptq_vs_qat}.
As the table shows, at all tested precisions QAT can grant a significant improvement w.r.t. PTQ. 
We appreciate the highest gap in the presence of activations quantized to low precision (4 bits), a setting which is particularly harmful for PTQ.
\input{tables/cityscapes_qat_vs_ptq.tex}
\paragraph{Semantic segmentation: VideoIQ}
We now aim to empirically compare \method to VideoIQ~\cite{sun2021dynamic} on a frame-prediction task, such as semantic segmentation on Cityscapes.
As VideoIQ is tailored for action recognition, it utilizes Temporal Segment Network~\cite{tsn} as a base model.
Therefore, we adapt it to use it along with segmentation architectures.
Specifically, we consider the QAT Frame Quantization models represented in \cref{fig:cityscapes_qat}, quantized at different precisions (8, 4, 3 and 2 bits).
We then train, following the objective in~\cite{sun2021dynamic}, a MobileNet v2~\cite{mobilenetv2} policy network that, given a frame, predicts the quantization level to be applied to the segmentation model.
By properly weighting the VideoIQ objectives, it is possible to obtain policies that reward BOP reduction and accuracy gains differently.
We represent the result of this VideoIQ model in \cref{fig:cityscapes_qat}.
Importantly, we advantage this model by excluding the cost of the (full precision) policy in the BOP count, and we only measure the strength of the dynamic selection of quantized models.
Even so, many times the policy degenerates to non-dynamic decisions, and fixes the predicted precision regardless of the input frame.
Moreover, whenever actually predicting bit-widths dynamically, the policy decisions do not outperform random selection (represented by dashed lines between models).
These results, consistent across four different backbones, suggest that a VideoIQ-like policy to dynamically select the quantization level based on the frame might be hard to learn for frame-prediction tasks.
\paragraph{Semantic segmentation: channel quantization.}
We investigate the benefits of performing channel quantization.
In \cref{tab:tensor_vs_channel_quantization}, we compare per-tensor and per-channel scale factors for PTQ \method on Cityscapes. 
We observe that channel quantization is on par or exceeds tensor quantization (up to $+1.78$ mIoU) for multiple precisions and is therefore favorable in \method.
\input{tables/cityscapes_tensor_vs_channel}
\paragraph{Semantic segmentation: per class analysis.}
\input{figures/cityscapes/cityscapes_per_class}
To further analyze the behaviour of residual quantization, we take a closer look at performances over different classes in Cityscapes.
Specifically, in \cref{fig:cityscapes_per_class} we measure the mIoU per class obtained with QAT for \basemethod and for ResQ, when the frame of interest is processed at 2 bits (for \method, we use \amprec{W3A3}{W2A2}).
We appreciate that \method typically performs better than the baseline on classes that have low mIoU (\ie, the challenging classes), such as rider, pole, traffic light and fence. 
We hypothesize that this behavior is due to the higher precision of the keyframe, serving as a good starting point for the residuals: 
applying a low precision update might be easier than detecting hard classes with a low precision from scratch.
\paragraph{Pose estimation: pairwise vs recurrence scheme.}
To further analyze the temporal aspects of \method, we consider an alternative sigma-delta decomposition scheme.
Specifically, instead of defining the residual as $\delta^t = x^t-x^k$ (pairwise), we might compute them as instantaneous variations, as $\delta^t = x^t-x^{t-1}$ (recurrent), as done in~\cite{skipconv,deltacnn}.
In this latter strategy, errors are likely to compound over time, as opposed to the pairwise strategy that only depends on the keyframe.
In \cref{fig:pairwise_vs_recurrent} we compare both summation strategies in \method, when applied to JHMDB (split 1).
Although for some precisions no significant difference can be noticed (\amprec{W8A8}{W4A8}), we appreciate how pairwise summations prove more stable over long intervals (\amprec{W8A8}{W8A4}).
We ascribe this finding to the error propagation hampering recurrent summations.
\input{figures/jhmdb_pairwise_vs_recurrent}

%% file: figures/cityscapes/combined_qat.tex
\begin{figure}[b]
\centering
\resizebox{\columnwidth}{!}{
\begin{tabular}{cc}
\includegraphics[width=0.25\textwidth]{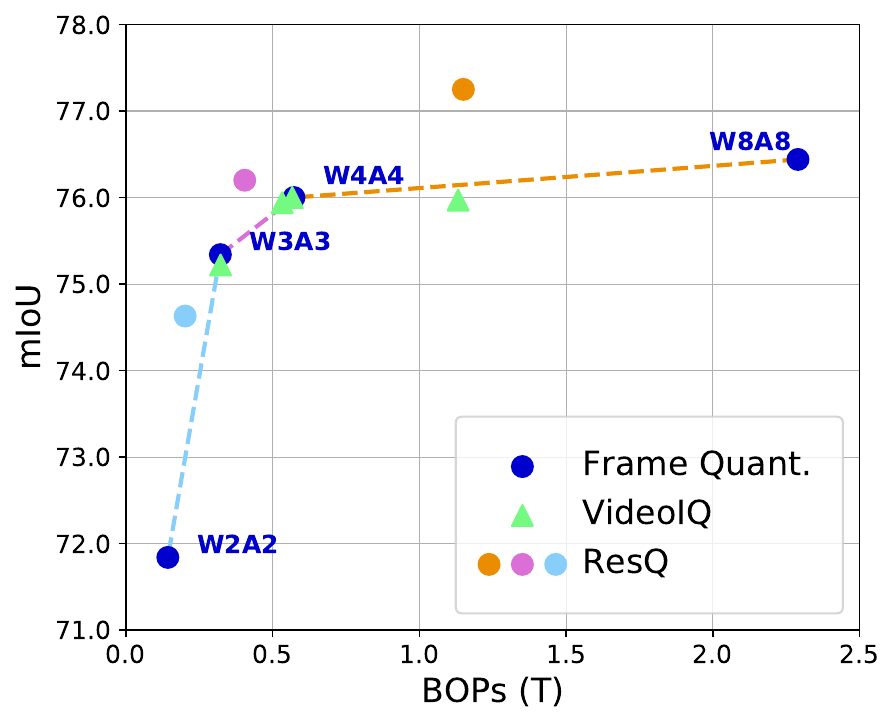}&
\includegraphics[width=0.25\textwidth]{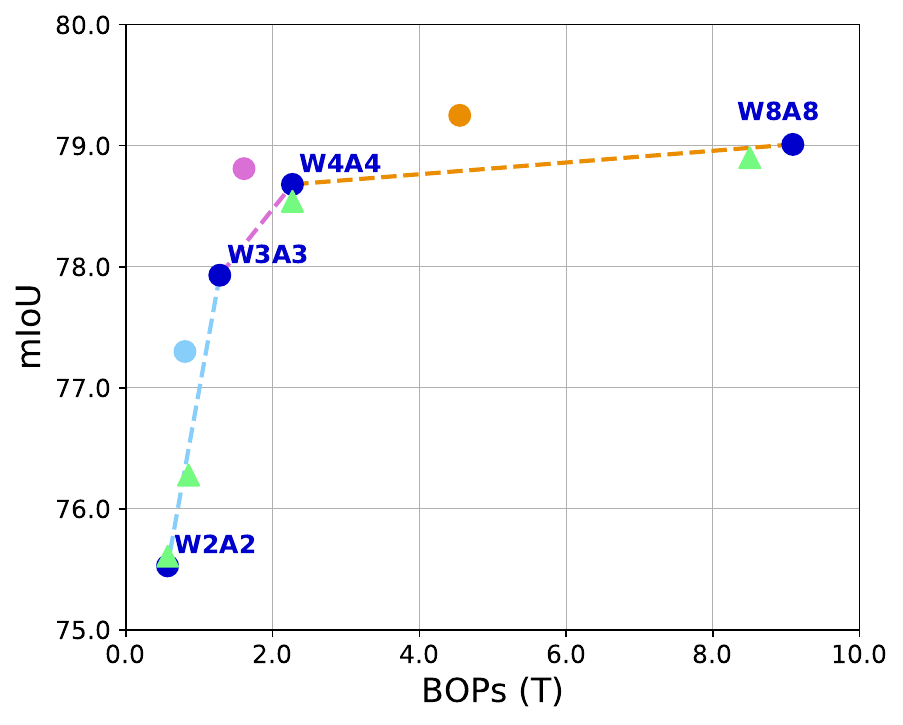}\\
(a) DDRNet23 slim & (b) DDRNet23 \\
\includegraphics[width=0.25\textwidth]{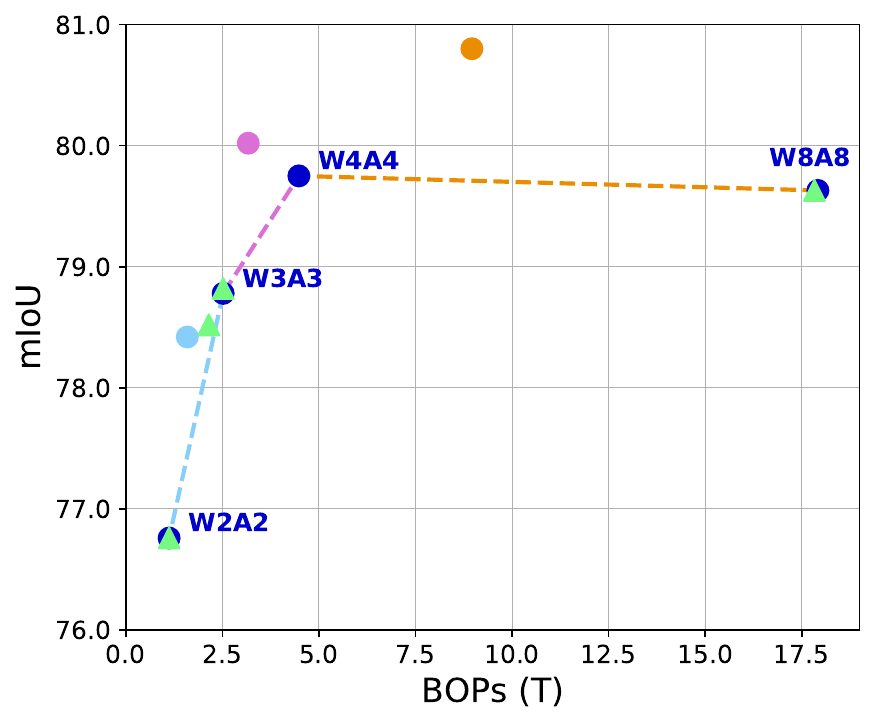}&
\includegraphics[width=0.25\textwidth]{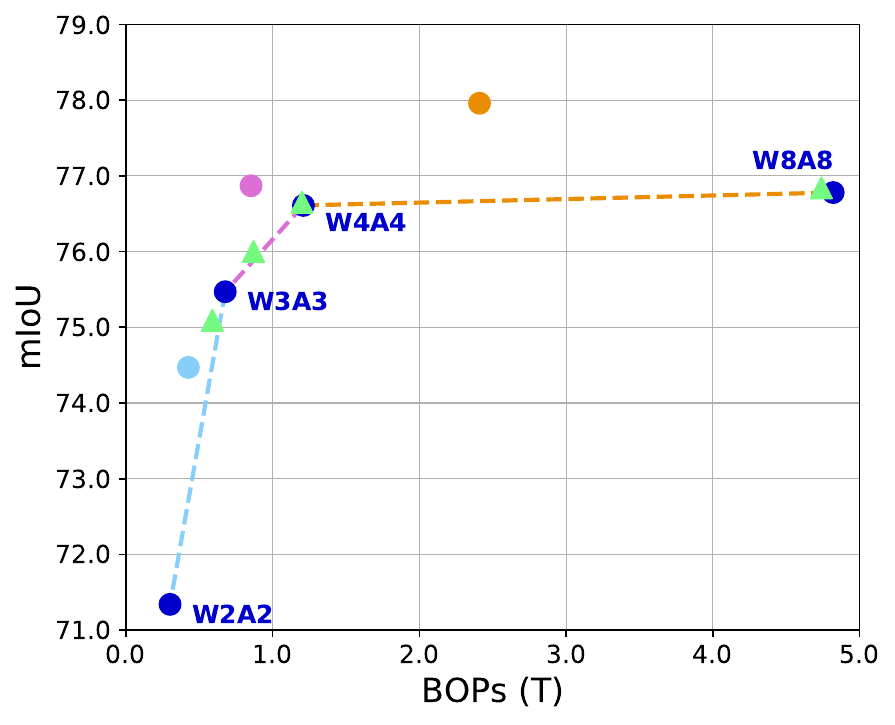}\\
(c) DDRNet39 & (d) HRNetw18-small
\end{tabular}}
\caption{\textbf{QAT results} for different DDRNet and HRNet models with \basemethod in blue and \method otherwise. Color specifies the operating precision between keyframes and residual frames, following dashed lines. Green triangles represent VideoIQ results.}
\label{fig:cityscapes_qat}
\end{figure}

%% file: tables/cityscapes_qat_vs_ptq.tex
\begin{table}[t]
\begin{center}
\resizebox{0.80\columnwidth}{!}{
\begin{tabular}{lccc}
\toprule
\textbf{Bit-width} & \textbf{Quantization} & \textbf{mIoU}& \textbf{$\Delta$ mIoU}\\
\midrule
\multirow{2}{*}{\amprec{W8A8}{W4A8}}& PTQ & 75.14  & - \tinyspace\\
& QAT &  \textbf{77.81}   & \textbf{+2.67}  \tinyspace\\
\midrule
\multirow{2}{*}{\amprec{W8A8}{W8A4}} & PTQ & 66.76 & - \tinyspace\\
& QAT & \textbf{77.70} & \textbf{+10.94}  \tinyspace\\
\midrule
 \multirow{2}{*}{\amprec{W8A8}{W4A4}} & PTQ & 66.67  & - \tinyspace\\
& QAT & \textbf{77.25} & \textbf{+10.58}  \tinyspace\\
\bottomrule
\end{tabular}}
\end{center}
\caption{\textbf{PTQ vs. QAT.} \method results on Cityscapes using DDRNet23-small at various precisions. Note that QAT can recover performance for precisions that are challenging for PTQ.}
\vspace{-1em}
\label{tab:ptq_vs_qat} 
\end{table}

%% file: tables/cityscapes_tensor_vs_channel.tex
\begin{table}[t]
\begin{center}
\resizebox{0.80\columnwidth}{!}{
\begin{tabular}{lccc}
\toprule
\textbf{Scheme} &  \textbf{Bit-width}    &  \textbf{mIoU}$\uparrow$  &  \textbf{$\Delta$ mIoU} $\uparrow$ \\
\midrule
\rowcolor{Gray}
 Tensor &  \amprec{W8A8}{W4A8}   &  73.41 & - \tinyspace\\
 Channel &  \amprec{W8A8}{W4A8}   &  75.14 & + \textbf{1.73} \tinyspace\\
\rowcolor{Gray}
 Tensor &  \amprec{W8A8}{W8A4}   &  66.59 & - \tinyspace\\
 Channel &  \amprec{W8A8}{W8A4}   &  66.76 & + 0.08 \tinyspace\\
 \rowcolor{Gray}
 Tensor &  \amprec{W8A8}{W4A4}   &  66.51 & - \tinyspace\\
 Channel &  \amprec{W8A8}{W4A4}   &  66.67 & + 0.16 \tinyspace\\
\rowcolor{Gray}
 Tensor &  \amprec{W8A4}{W4A4}   &  58.43 & - \tinyspace\\
 Channel &  \amprec{W8A4}{W4A4}   &  60.21 & + \textbf{1.78} \tinyspace\\
\midrule
\end{tabular}}
\end{center}
\caption{\textbf{Ablation on Tensor vs. Channel Quantization.} Experiments on \method at various bit-widths.}
\label{tab:tensor_vs_channel_quantization} 
\end{table}

%% file: figures/cityscapes/cityscapes_per_class.tex
\begin{figure}
\centering
\includegraphics[width=0.9\columnwidth]{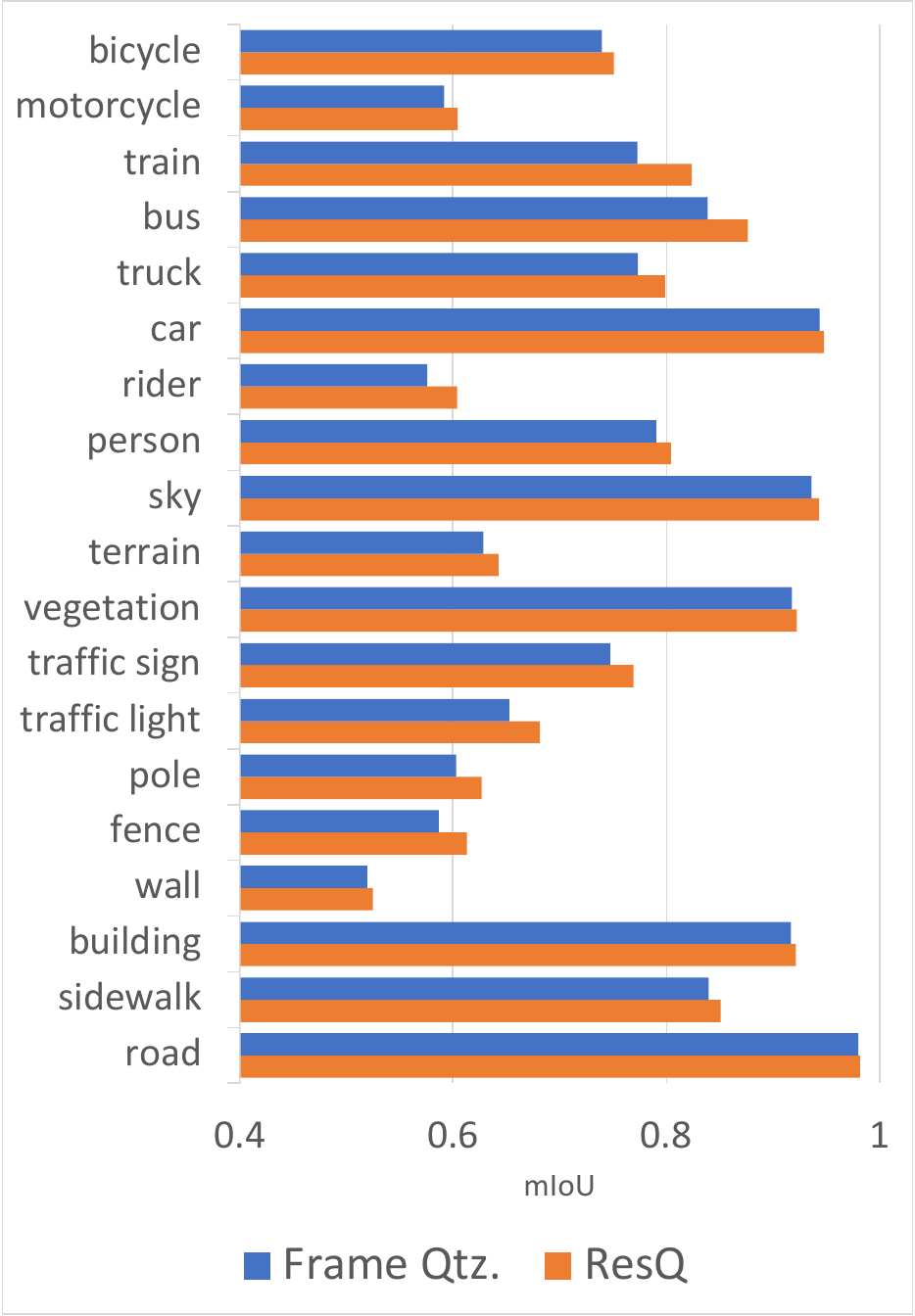}\\
\caption{\textbf{mIoU per class} on Cityscapes for \basemethod and \method when the frame of interest is quantized at 2 bits.}
\label{fig:cityscapes_per_class}
\end{figure}

%% file: figures/jhmdb_pairwise_vs_recurrent.tex
\begin{figure}[t]
\centering
\includegraphics[width=0.9\columnwidth]{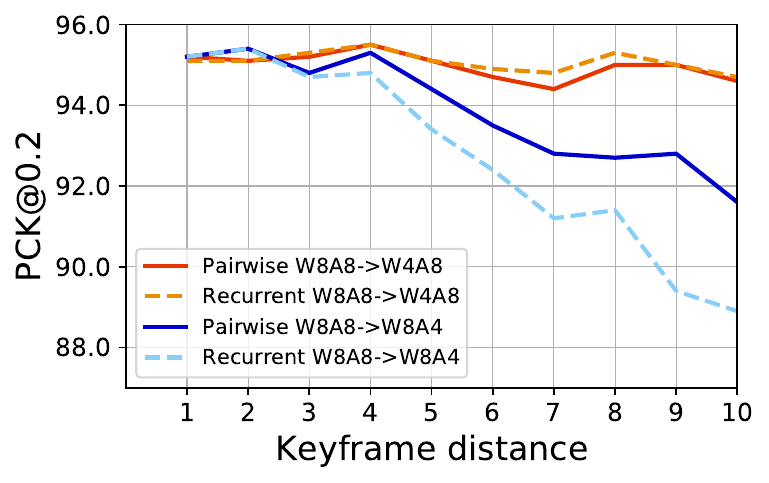}
\caption{\textbf{Ablation on pairwise vs. recurrent} summation on JHMDB for different bit-widths. Note that activation quantization suffers from using recurrent summation and pairwise is better or equal as we increase the distance to the keyframe.}
\label{fig:pairwise_vs_recurrent}
\end{figure}

%% file: sections/supp_text/latency.tex
\section{BOP and inference time}
To investigate whether theoretical BOPs gains would translate to actual runtime improvements that processing at low bit-widths would bring, we rely on a HW simulator for future generations of SnapDragon NPU, a low-power fixed-point accelerator.
\input{tables/inference_simulation.tex}
Tab.~\ref{tab:inference_simulation} shows how bringing weights and activations from 8 to 4 bits reduces the simulated runtime for both pose-estimation and semantic segmentation models.

%% file: tables/inference_simulation.tex
\begin{table}[t]
\begin{center}
\resizebox{0.9\columnwidth}{!}{
\begin{tabular}{lcccc}
\toprule
\textbf{Model} & \textbf{W8A8}  &  \textbf{W4A8} & \textbf{W8A4} & \textbf{W4A4}\\ 
\midrule
DDRNet-23s &  3.52 & 2.76 & 2.43 &  1.96\\ 
\midrule
 HRNet-w32 &   0.93 & 0.68 & 0.72 & 0.49\\ 
\midrule
\end{tabular}
}
\end{center}
\vspace{-1em}
\caption{\textbf{Simulated runtimes} in milliseconds for semantic segmentation (DDRNet-23s) and pose estimation (HRNet-w32).}
\label{tab:inference_simulation} 
\end{table}

%% file: sections/supp_text/qualitative_results.tex
\section{Qualitative results}
We report additional qualitative comparison between frame and residual quantization in \cref{fig:cityscapes_qualitative_supplementary}.
We report the comparison at various bit-widths (4, 3, and 2 bits).
Similarly to the example reported in Fig.~8 in the main paper, in highlighted regions \method takes advantage of the high precision representation granted by the keyframe to better segment the scene at hand with low precision.
\input{figures/cityscapes/cityscapes_qualitative_supplementary}

%% file: figures/cityscapes/cityscapes_qualitative_supplementary.tex
\bgroup
\setlength{\tabcolsep}{1pt}
\begin{figure*}[t]
\centering
\includegraphics[width=\textwidth]{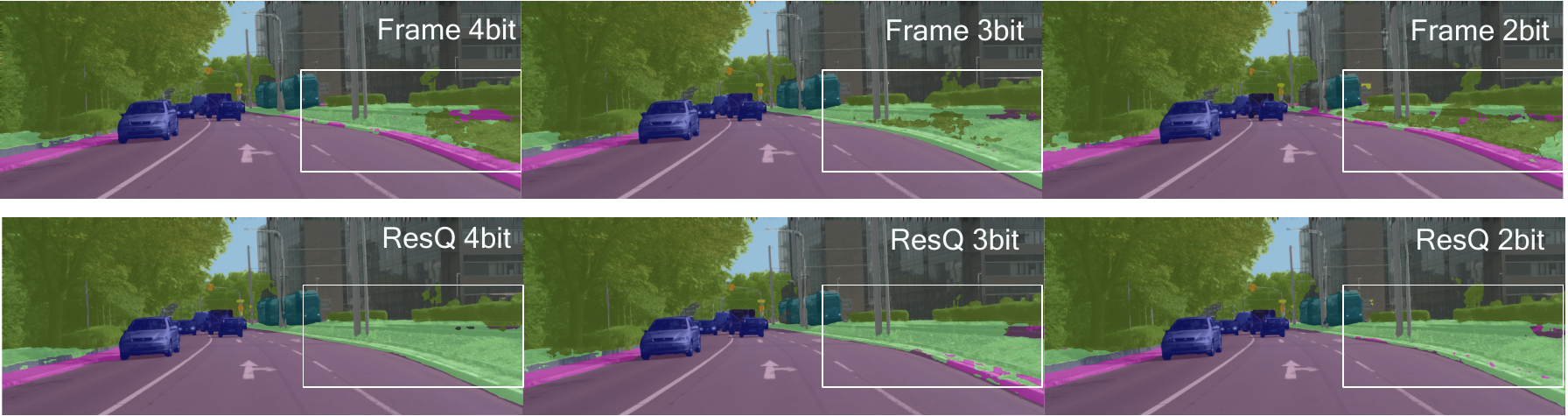} \\
\includegraphics[width=\textwidth]{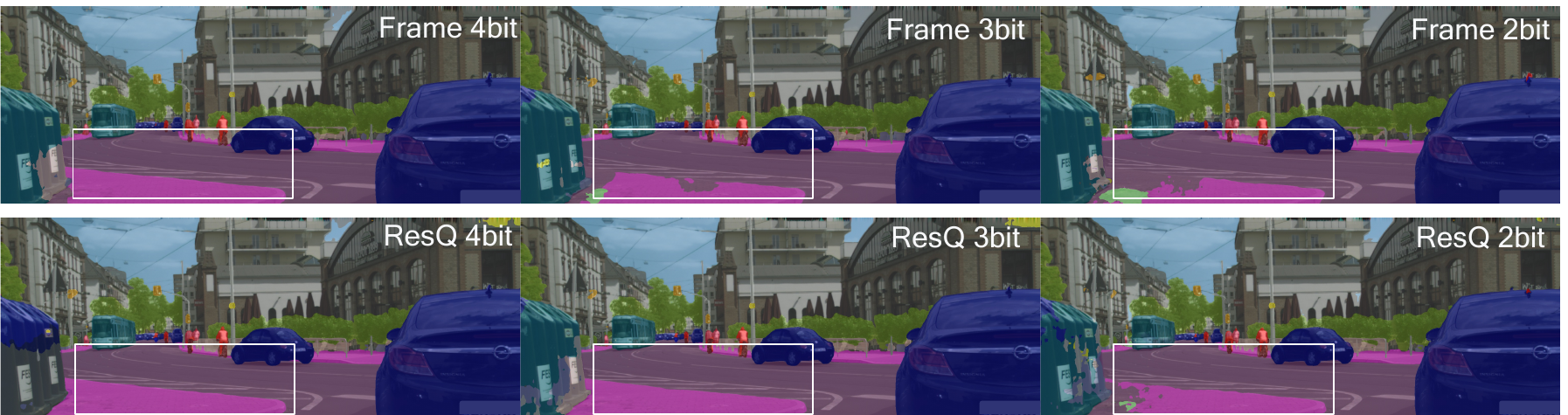} \\
\includegraphics[width=\textwidth]{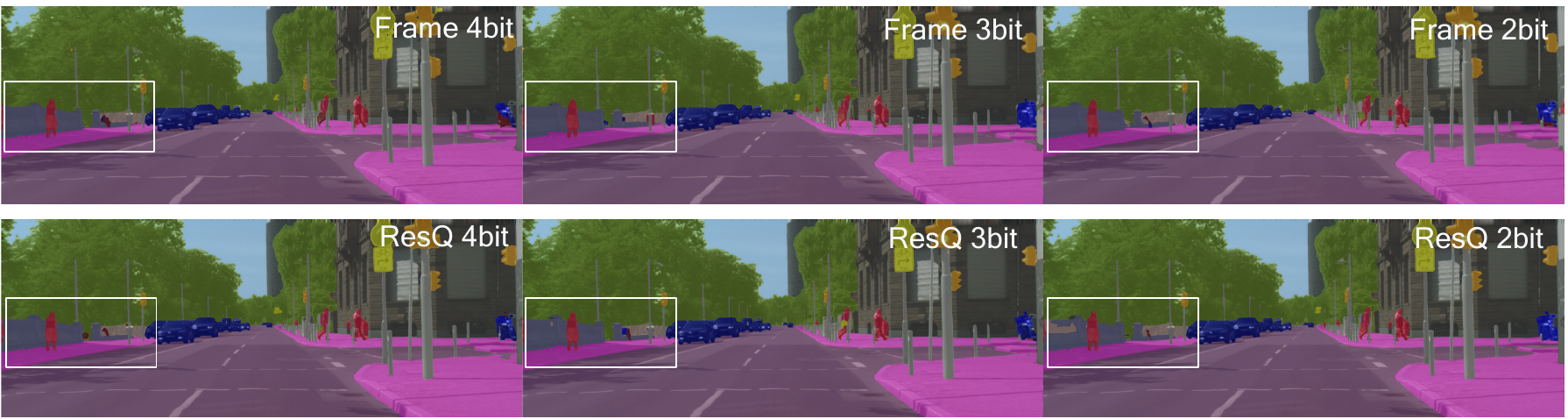} \\
\caption{\textbf{Qualitative comparison} between frame and residual quantization at low bit-width. All models illustrated here benefit of QAT. In highlighted regions, our proposal significantly outperforms the baseline.}
\label{fig:cityscapes_qualitative_supplementary}
\end{figure*}
\egroup